\documentclass[10pt,onecolumn]{article}

\usepackage[margin=0.75in]{geometry}
\usepackage{float}
\usepackage{times}
\usepackage{amsmath}
\usepackage{amsfonts}
\usepackage{booktabs}
\usepackage{array}
\usepackage{longtable}
\usepackage{url}
\usepackage{graphicx}
\usepackage{caption}
\usepackage{balance}
\usepackage{natbib}

\usepackage{listings}
\usepackage{xcolor}
\definecolor{codegreen}{rgb}{0,0.6,0}
\definecolor{codegray}{rgb}{0.5,0.5,0.5}
\definecolor{codepurple}{rgb}{0.58,0,0.82}

\lstdefinelanguage{yaml}{
  keywords={true,false,null,yes,no},
  keywordstyle=\color{codepurple}\bfseries,
  sensitive=false,
  comment=[l]{\#},
  commentstyle=\color{codegreen}\ttfamily,
  stringstyle=\color{codepurple}\ttfamily,
  morestring=[b]',
  morestring=[b]"
}

\lstdefinelanguage{json}{
    basicstyle=\small\ttfamily,
    numbers=left,
    numberstyle=\tiny,
    stepnumber=1,
    numbersep=8pt,
    showstringspaces=false,
    breaklines=true,
    frame=lines,
    string=[s]{"}{"},
    stringstyle=\color{blue},
    comment=[l]{:},
    commentstyle=\color{black},
}

\usepackage{titling}
\pretitle{\begin{center}\LARGE\bfseries}
\posttitle{\par\end{center}\vskip 0.5em}
\preauthor{\begin{center}\large}
\postauthor{\end{center}}
\predate{\begin{center}\large}
\postdate{\par\end{center}\vskip 1em}

\usepackage{titlesec}
\titleformat{\section}{\large\bfseries}{\thesection}{1em}{}
\titleformat{\subsection}{\normalsize\bfseries}{\thesubsection}{1em}{}
\titleformat{\subsubsection}{\normalsize\bfseries}{\thesubsubsection}{1em}{}

\renewenvironment{abstract}
{\small\quotation\noindent\textbf{Abstract}\par}
{\endquotation}

\begin{document}

\title{Scalable and Reliable Evaluation of AI Knowledge Retrieval Systems: RIKER and the Coherent Simulated Universe}

% Custom column type for tables
\newcolumntype{L}[1]{>{\raggedright\arraybackslash}p{#1}}
\newcolumntype{C}[1]{>{\centering\arraybackslash}p{#1}}

\author{JV Roig\\
\small Kamiwaza AI\\
\small \texttt{jv@kamiwaza.ai}
}

\date{December 2025}

\maketitle

\begin{abstract}
\noindent
Evaluating knowledge systems (LLMs, RAG, knowledge graphs, etc) faces fundamental challenges: static benchmarks are vulnerable to contamination, LLM-based judges exhibit systematic biases, and ground truth extraction requires expensive human annotation. We present RIKER (Retrieval Intelligence and Knowledge Extraction Rating), both a benchmark and a replicable methodology based on paradigm inversion - generating documents \textit{from} known ground truth rather than extracting ground truth \textit{from} documents. This approach enables deterministic scoring and scalable evaluation without human annotation or reference models, and contamination resistance through regenerable corpora. Our evaluation of 33 models using over 21 billion tokens reveals that context length claims frequently exceed usable capacity, with significant degradation beyond 32K tokens; cross-document aggregation proves substantially harder than single-document extraction; and grounding ability and hallucination resistance are distinct capabilities - models excelling at finding facts that exist may still fabricate facts that do not. Beyond the specific benchmark, we contribute a domain-agnostic methodology for constructing scalable and contamination-resistant evaluations wherever synthetic documents can be generated from structured ground truth.
\end{abstract}

\begin{figure}[H]
\centering
\includegraphics[width=\textwidth]{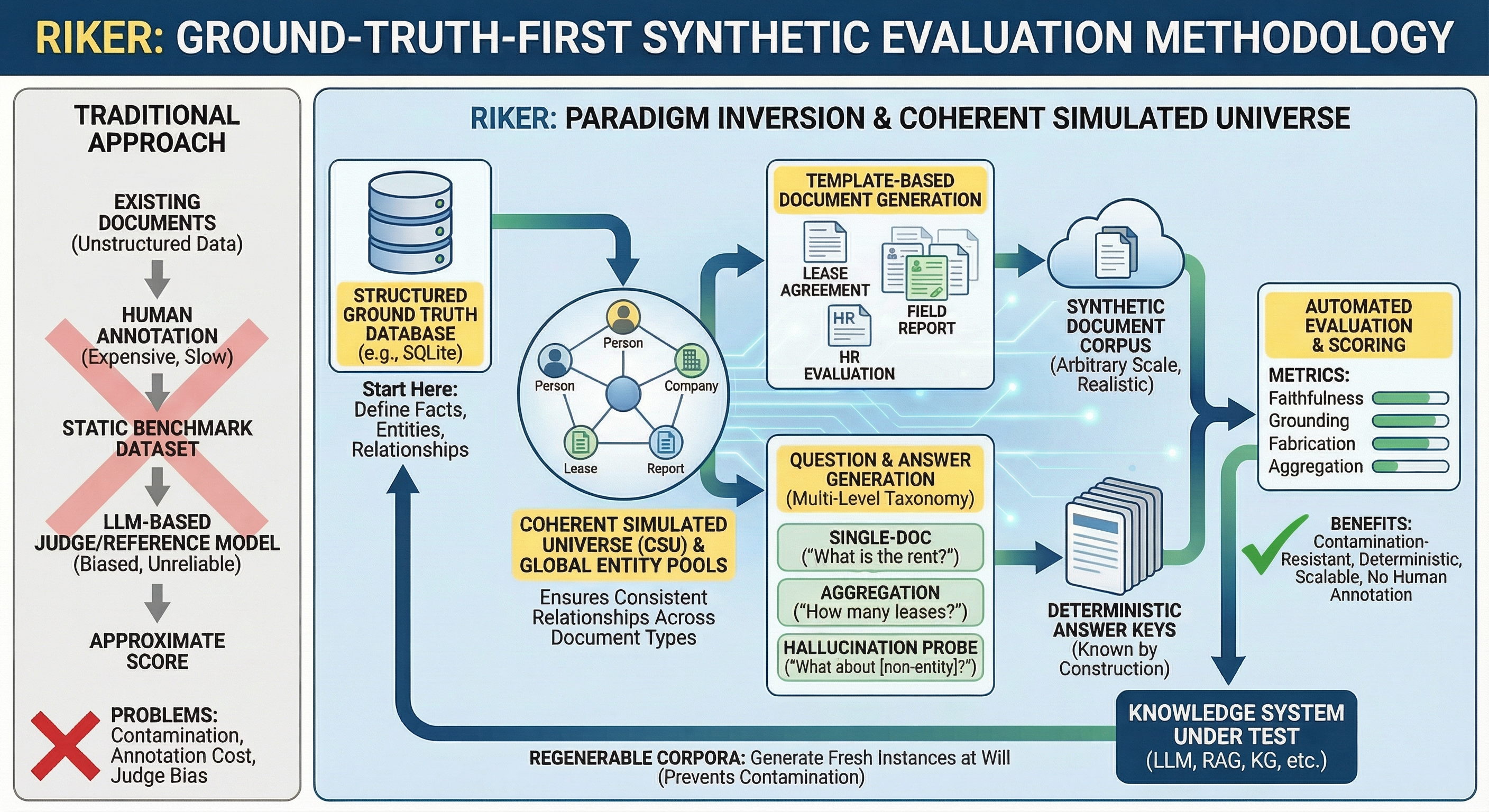}
\caption{RIKER methodology overview. Traditional approaches (left) extract ground truth from existing documents via expensive human annotation, producing static benchmarks vulnerable to contamination and requiring biased LLM judges. RIKER (right) inverts this: structured ground truth is defined first, then documents and questions are generated from it, enabling deterministic scoring and regenerable corpora.}
\label{fig:riker-overview}
\end{figure}

\section{Introduction}\label{sec:intro}

One of the most common and also most critical uses of agentic AI in the enterprise is processing vast amounts of internal enterprise knowledge. This comes in many forms:
\begin{itemize}
    \item Relevant enterprise documents are loaded into the context of the large language model (LLM) at the beginning of a chat session
    \item Snippets of relevant enterprise documents are loaded as needed, through a knowledge retrieval mechanism - such as through various Retrieval Augmented Generation (RAG) methodologies using vector databases, traditional enterprise search, ontologies, or any mix of these techniques
    \item An LLM agent, given appropriate tools and access, can also retrieve internal and external information as needed, in an Agentic RAG manner.
\end{itemize}

The list above is not meant to be exhaustive - merely illustrative of the many ways that LLMs are deployed to be able to use enterprise knowledge in order to be useful to the enterprise.

The huge enterpise gap here is: how do we QA (quality assurance) all of these in a scalable and reliable manner? This simple question extends to many related questions that enterprise teams can find themselves asking: \textit{Which model hallucinates less? Which model is better retrieving facts if we dump documents into its context? Should I use a vector database or a graph database? How do I test and quantify how much better or worse a chunking/embedding/retrieval configuration is better over another? How much does an ontology improve our knowledge retrieval?}

As before, the list above is not meant to be exhaustive, just illustrative of the gap. This gap exists because, currently, answering questions like these is extremely difficult.

If this gap can be solved, it will not only be useful to enterprise, but also to the research community. A tool that can quantify how accurate or inaccurate certain retrieval methods are (from simple LLM context stuffing, to ornate ontologies and knowledge graphs) would be an extremely useful tool not only as a primary QA tool in the enterprise, but could help the research and development of improved retrieval systems.

In this work, we propose one such tool - Retrieval Intelligence and Knowledge Extraction Rating (RIKER). RIKER is an offshoot of our earlier work called PICARD (Probing Intelligent Capabilities via Artificial Randomized Data) \cite{roig2025picard} - a framework for contamination-resistant benchmarking of agentic AI capabilities of LLMs. RIKER builds upon PICARD to extend evaluation to knowledge extraction scenarios, as well as evaluation of knowledge retrieval systems in general beyond just LLMs.

We present RIKER as two contributions: first, a concrete benchmark for enterprise document understanding with empirical results across 31 models; second, and more broadly, a replicable \textit{methodology} for constructing scalable and reliable evaluations of knowledge retrieval systems. The methodology - generating documents FROM known ground truth rather than extracting ground truth FROM documents - is domain-agnostic and can be applied wherever synthetic documents can be generated from structured facts. Figure~\ref{fig:riker-overview} illustrates the approach.

The key contributions of this work are:
\begin{itemize}
    \item \textbf{Ground truth by construction} through paradigm inversion - rather than extracting ground truth from documents, RIKER generates documents FROM known ground truth, eliminating human annotation
    \item \textbf{Scalable and reliable benchmarking} through procedural generation - regenerable corpora and tests create fresh, never-before-seen documents and questions. Documents and tests are independently scalable.
    \item The \textbf{Coherent Simulated Universe} approach - synthetic documents that maintain realistic entity relationships across document types
    \item A \textbf{multi-level question taxonomy} spanning single-document extraction, cross-document aggregation, and hallucination detection
    \item \textbf{Empirical evaluation of 31 models} across 32K, 128K, and 200K token contexts, totaling nearly 20 billion tokens of evaluation
\end{itemize}

Our evaluation reveals several key findings: top-tier models achieve over 80\% accuracy at 32K context but degrade significantly at longer contexts; aggregation queries prove substantially harder than single-document extraction; and several models exhibit catastrophic failure modes including coherence loss and hallucination spikes exceeding 70\%. Cross-corpus validation with independent document sets confirms that RIKER measures model capability rather than corpus-specific artifacts.

\section{Related Work}

Evaluating knowledge retrieval and extraction systems faces fundamental methodological challenges. Current approaches rely on static benchmarks vulnerable to contamination, LLM-based judges with documented biases, and approximated ground truth derived from expensive human annotation. This section reviews the current state of knowledge retrieval evaluations of LLMs and various knowledge retrieval systems like RAG and knowledge graphs.

\subsection{Long-Context LLM Evaluation}

The Needle-in-a-Haystack (NIAH) paradigm \cite{kamradt2023niah} places a random fact in a long context and tests retrieval. While influential, the original NIAH fundamentally tests \textit{retrieval}, not \textit{comprehension} - finding an arbitrary fact buried in unrelated text is pattern-matching, not document understanding. Ironically, the needle's incongruence with its context makes it \textit{easier} to find; it stands out precisely because it doesn't belong.

Subsequent benchmarks address various limitations with meaningful improvements. \textbf{Sequential-NIAH} \cite{yu2025sequentialniahneedleinahaystackbenchmarkextracting} tests whether models can extract facts in correct chronological order across 8K - 128K contexts, demonstrating that sequential understanding remains challenging even for frontier models. \textbf{RULER} \cite{hsieh2024rulerwhatsrealcontext} expands beyond retrieval to include multi-hop tracing, aggregation, and question answering tasks; despite models claiming 32K+ context support, many fail to maintain satisfactory performance at that length. \textbf{NeedleChain} \cite{moon2025needlechainmeasuringintactlongcontext} takes a stricter approach: every piece of context is essential for answering queries, so missing even one element results in failure - revealing that models achieving near-perfect NIAH scores struggle under these stricter conditions.

For more realistic evaluation, \textbf{LongBench v2} \cite{bai2025longbenchv2deeperunderstanding} provides human-annotated tasks across six categories (single/multi-document QA, code understanding, dialogue history, structured data) with contexts from 8K to 2M words, demonstrating that long-context comprehension remains challenging even for frontier models. \textbf{InfiniteBench} \cite{zhang2024inftybenchextendinglongcontext} pushes context length to 100K+ tokens using real content from novels, code repositories, and mathematical problems.

Additional benchmarks serve specialized purposes: U-NIAH \cite{gao2025uniahunifiedragllm} provides unified comparison between RAG and LLM approaches on NIAH-style tasks, while MMNeedle \cite{wang2025multimodalneedlehaystackbenchmarking} extends the paradigm to multimodal (image) contexts.

The ``Lost in the Middle'' phenomenon \cite{liu2024lostinthemiddle} demonstrates that LLMs struggle with information placed in the middle of long contexts, with some mitigation approaches proposed \cite{zhang2024middlelanguagemodelsuse}.

\subsection{Retrieval and Embedding Benchmarks}

BEIR \cite{thakur2021beir} and MTEB \cite{muennighoff2023mtebmassivetextembedding} are the standard benchmarks for retrieval evaluation. Newer domain-specific benchmarks include FreshStack \cite{freshstack2025} for technical documents and MIRAGE \cite{xiong2024benchmarkingretrievalaugmentedgenerationmedicine} for medical retrieval. However, all face contamination concerns. BEIR is no longer a true zero-shot benchmark, as researchers now routinely include BEIR datasets in their training pipelines \cite{husain2024modernirevals}. MTEB's leaderboard now has 400+ models with marginal performance differences, suggesting either saturation or overfitting to the benchmark distribution.

FreshStack is less prone to contamination, but relies on LLM-as-a-judge for evaluation.

\subsection{Multi-Hop Question Answering}

Multi-hop QA benchmarks like HotpotQA \cite{yang2018hotpotqa} and 2WikiMultiHopQA \cite{ho2020wikimultihop} aim to test reasoning across multiple evidence pieces. More recent efforts include MoreHopQA \cite{schnitzler2024morehopqa} for deeper reasoning chains and multi-hop RAG approaches \cite{tang2024multihopragbenchmarkingretrievalaugmentedgeneration}. However, shortcut exploitation undermines validity: nearly 61\% of HotpotQA's multi-hop questions can actually be answered using single-hop reasoning, with a simple BERT-based single-hop model achieving performance comparable to state-of-the-art multi-hop systems \cite{min2019compositional}.

MuSiQue \cite{trivedi2022musique} was designed specifically to prevent shortcuts through ``unanswerable'' questions, but remains static and thus vulnerable to contamination.

\subsection{RAG Evaluation}

RAG evaluation has received significant survey attention \cite{yu2024evaluationretrievalaugmentedgenerationsurvey, gan2025retrievalaugmentedgenerationevaluation}. Popular frameworks like RAGAS \cite{es2024ragas} and ARES \cite{saadfalcon2024ares} rely on LLM-as-judge approaches, with domain-specific validation in areas like telecommunications \cite{telecomrag2024}. However, empirical validation reveals significant limitations: correlation between RAGAS metrics and human evaluation yields a harmonic mean of only 0.55, far below what would be required for reliable automated evaluation \cite{beatrust2024ragasevaluation}.

The CALM framework \cite{ye2024justiceprejudicequantifyingbiases} documents 12 distinct biases in LLM judges, including position bias \cite{shi2025judgingjudgessystematicstudy}, verbosity bias, self-enhancement bias, and authority bias. Additional critiques appear in \cite{gao2025evaluatingmitigatingllmasajudgebias}, \cite{cip2024llmjudgesunreliable}, and comparative studies of human vs.\ LLM judges \cite{chen2024humansllmsjudgestudy}. RAGChecker \cite{ru2024ragcheckerfinegrainedframeworkdiagnosing} advances the field with fine-grained diagnostic metrics but still uses LLM-based entailment checking.

For agentic RAG, evaluation remains nascent. Recent surveys \cite{liang2025reasoningrag12} and benchmarks like RAGCap-Bench \cite{lin2025ragcapbenchbenchmarkingcapabilitiesllms}, HopRAG \cite{liu2025hopragmultihopreasoninglogicaware}, and TelAgentBench \cite{telagentbench2025} address this gap, but RAGCap-Bench finds that current systems still struggle with challenging multi-hop questions and their intermediate reasoning capabilities remain underexplored. Community resources track ongoing developments \cite{awesomeragreasoning2025}.

\subsection{GraphRAG and Knowledge Graph Evaluation}

GraphRAG promises improved retrieval through knowledge graph augmentation \cite{awesomegraphrag2024}, but evaluation challenges persist. Benchmarks like GraphRAG-Bench \cite{graphragbench2025} and evaluation frameworks \cite{zeng2025significantrealperformancegains, sansford2024graphevalknowledgegraphbasedllm} attempt to standardize assessment, while studies examine when graphs help RAG \cite{xiang2025usegraphsragcomprehensive}. When evaluated with ground truth rather than LLM-as-judge, community-based GraphRAG, particularly with global search, generally underperforms compared to standard RAG \cite{han2025ragvsgraphragsystematic}. This discrepancy arises because the original GraphRAG evaluation \cite{edge2024graphrag} used LLM-as-judge without ground truth - precisely the methodology shown to inflate results.

Knowledge graph construction faces fundamental evaluation challenges: recent surveys identify unresolved issues in establishing reliable benchmarks for KG quality assessment, particularly regarding intrinsic and extrinsic evaluation metrics \cite{kgconstruction2025survey}. The relationship between KGs and hallucination has been studied extensively \cite{pan2024kghallucination}. GOSyBench \cite{gosybench2024} provides domain-specific KG extraction benchmarks, demonstrating that even frontier models struggle with accurate knowledge graph recovery. FinReflectKG \cite{finreflectkg2025} combines rule-based checks, statistical validation, and LLM-as-judge assessments to measure extraction quality - a multi-layered approach reflecting the difficulty of establishing absolute quality metrics for KG construction.

\subsection{Hallucination and Factuality Benchmarks}

Hallucination detection and factuality evaluation have received extensive attention \cite{huggingface2024hallucinations}. TruthfulQA and SimpleQA \cite{wei2024measuringshortformfactualitylarge} evaluate factual accuracy, but face distinct challenges. TruthfulQA shows evidence of contamination \cite{deng2024investigatingdatacontaminationmodern}, while SimpleQA focuses on short-form responses where even frontier models struggle to achieve majority accuracy. HaluEval \cite{li2023halueval} and HalluLens \cite{bang2025hallulensllmhallucinationbenchmark} provide additional hallucination benchmarks, with FastFact \cite{wan2025fastfact} offering efficient fact verification.

For long-form factuality, FActScore \cite{min2023factscore} decomposes responses into atomic facts for verification. Studies on generalization vs.\ memorization \cite{dong2024generalizationmemorization} inform understanding of when models hallucinate. No benchmark adequately addresses long-form knowledge extraction from document corpora.

\subsection{Benchmark Contamination}

Data contamination undermines benchmark validity across domains, with comprehensive surveys documenting the extent of the problem \cite{xu2024benchmarkdatacontaminationlarge, awesomedatacontamination2024}. Meta-analyses question benchmark trustworthiness broadly \cite{eriksson2025aibenchmarks}. Simple variations such as paraphrasing or translation easily bypass standard decontamination measures, allowing a 13B model to overfit leaked benchmarks and achieve GPT-4-level performance \cite{yang2023rephrasedsamples}.

MMLU shows 52 - 57\% exact-match guessing rates on contaminated subsets \cite{deng2024investigatingdatacontaminationmodern}. Current solutions - LatestEval \cite{li2024latesteval} (temporal freshness), MMLU-CF \cite{zhao2024mmlucfcontaminationfreemultitasklanguage} (counterfactual rephrasing) - are \textit{reactive}: they detect and mitigate contamination rather than prevent it structurally.

\subsection{Synthetic Data Generation and Simulation Validity}

RIKER generates synthetic documents from structured ground truth, situating it within the synthetic data literature \cite{awesomellmsyntheticdata2024}. Surveys on LLM-driven synthetic data generation \cite{long2024llmsdrivensyntheticdatageneration} and user simulation \cite{usersimulation2025} provide theoretical grounding.

\subsubsection{The Bias Factor Problem}

When LLMs generate benchmark data and perform the task, systematic biases emerge. Smaller LLMs exhibit biases towards their own generated data, whereas larger models do not - a phenomenon termed the ``bias factor'' \cite{syntheticbenchmarkefficacy2024}. This undermines validity for LLM-generated benchmarks. RIKER avoids this through \textbf{template-based generation} - no LLM is involved in document creation.

\subsubsection{The Reality Gap}

The ``reality gap'' - the gap between synthetic training data and real-world deployment data - is not merely technical but epistemological: synthetic data requires real-world data about a domain in order to model it, the very data it purports to dispense with \cite{steinhoff2025realitygap}. A validity-centered framework for AI evaluation \cite{validity2025framework} provides psychometric grounding for what claims benchmarks can legitimately support. RIKER makes bounded validity claims: it measures extraction accuracy against known ground truth. It does \textit{not} claim that success on RIKER guarantees real-world performance. Rather, RIKER provides a \textit{necessary} (not sufficient) condition - if a system cannot extract known facts from synthetic documents, it will fail on real documents.

\subsubsection{Parameterized Evaluation}

Two benchmarks validate template-based evaluation:

\textbf{GSM-Symbolic} \cite{mirzadeh2024gsmsymbolic} uses symbolic templates for math problems, finding that adding irrelevant clauses causes up to 65\% performance drops. The authors conclude: ``Current LLMs cannot perform genuine logical reasoning; they replicate reasoning steps from their training data.''

\textbf{RV-Bench} \cite{hong2025rvbench} generates Random Variable Questions (RVQs) with randomized variable combinations, testing 30+ LLMs and finding ``proficiency imbalance'' between familiar and novel combinations.

Both GSM-Symbolic and RV-Bench demonstrate that parameterized generation exposes capability gaps that static benchmarks miss. RIKER extends this paradigm from mathematical reasoning to knowledge extraction.

\subsection{The PICARD Framework}

The PICARD framework \cite{roig2025picard} addresses evaluation gaps for agentic AI through:

\begin{itemize}
    \item \textbf{Ground-truth-first generation}: ``When the evaluation framework controls data generation, it inherently possesses complete ground truth''
    \item \textbf{Deterministic scoring}: Answer keys generated simultaneously with test data
    \item \textbf{Anti-memorization by design}: Combinatorial explosion makes memorization impossible
    \item \textbf{Multi-layered randomization}: Entity substitution, data generation, and environmental variation
\end{itemize}

PICARD demonstrates these principles for file manipulation, database operations, and multi-step workflows. However, PICARD does not address knowledge extraction from document corpora - the domain RIKER targets.

\subsection{Summary: Research Gaps}

Table~\ref{tab:gap-analysis} summarizes the gaps addressed by RIKER.

\begin{longtable}{@{}L{4.2cm}L{4.2cm}L{4.2cm}@{}}
\caption{Research Gaps and RIKER's Position} \label{tab:gap-analysis} \\
\toprule
\textbf{Gap} & \textbf{Current Limitation} & \textbf{RIKER's Approach} \\
\midrule
\endfirsthead
\toprule
\textbf{Gap} & \textbf{Current Limitation} & \textbf{RIKER's Approach} \\
\midrule
\endhead
\bottomrule
\endfoot
\bottomrule
\endlastfoot

LLM-as-Judge unreliability & 12 documented biases; 0.55 human correlation & Deterministic scoring \\ \midrule
Benchmark contamination & Static benchmarks vulnerable to memorization & Regenerable corpora + combinatorial anti-memorization \\ \midrule
Long-context retrieval $\neq$ comprehension & Needle retrieval $\neq$ document comprehension & Realistic document understanding \\ \midrule
Static multi-hop QA & Single-hop shortcuts & Aggregation requires cross-document extraction \\ \midrule
RAG ground truth approximated & Human annotation bottleneck & Ground-truth-first generation \\ \midrule
No KG extraction ground truth & Reliable KG benchmark establishment remains unresolved & Generate FROM structured ground truth with known entity relationships \\ \midrule
KG completeness unmeasurable & Graph edit distance insufficient & SQLite manifest defines expected output \\ \midrule
Retrieval benchmark contamination & BEIR no longer zero-shot & Regenerable corpus \\ \midrule
KG evaluation schema-dependent & Tied to specific ontology & Query answers, not structure \\ \midrule
Data bottleneck & Annotation expensive, doesn't scale & Arbitrary-scale generation \\ \midrule
GraphRAG evaluation inflated & LLM-judge without ground truth & Ground truth scoring \\

\end{longtable}

The pattern reveals three fundamental problems: (1) static datasets get contaminated, (2) LLM judges are unreliable, and (3) ground truth is approximated rather than known. Industry reports confirm that data bottlenecks have increased significantly \cite{appen2024stateofai}, while best practices for ground truth generation remain labor-intensive \cite{aws2024groundtruth}. RIKER's paradigm inversion - generating documents FROM ground truth rather than extracting ground truth FROM documents - addresses all three.

\section{The RIKER Approach}

Unlike benchmarks that rely on static datasets or LLM-as-judge evaluation, RIKER generates synthetic corpora from embedded ground truth, enabling deterministic scoring at scale. The methodology is application-agnostic - it can evaluate context-stuffing, retrieval-augmented generation, or knowledge graph systems by instrumenting the system under test to answer RIKER-generated questions.

\subsection{Synthetic Corpus Generation}

RIKER employs a ground-truth-first architecture: the complete knowledge base—all entities, relationships, and facts—is populated in a relational database \textit{before} any document is generated. Documents are then rendered as human-readable views of this underlying ground truth. This inversion of the typical ``extract facts from documents''approach provides three key advantages: (1) every question has a verifiable answer by construction, enabling deterministic scoring,  (2) corpora can be regenerated with different random seeds while maintaining structural equivalence, enabling robustness validation, and (3) the approach is easily scalable to practically-unlimited scale, requiring no human-intensive document annotation.

\subsubsection{Synthetic Data Generation}
Built upon the PICARD framework \cite{roig2025picard}, RIKER inherits and expands various synthetic data generation functions - names, amounts, dates, and other entities - which are used to generate a diverse set of ground truth elements, from which documents will later be created.

\subsubsection{Ground Truth Database} \label{sec:GroundTruthDB}

All generated ground truth is recorded in a SQLite database with full relational structure. For example, for a lease document corpus, this includes:
\begin{itemize}
    \item Document metadata (parties, dates, amounts, clauses present, clauses absent)
    \item Entities (lessors, lessees, agents, addresses, etc.)
    \item Entity relationships (which lessors have which lessees)
\end{itemize}

This database serves as the authoritative answer key for all generated questions. Questions that require computed aggregations (counts, sums, temporal relationships) can be derived from this ground truth through SQL, which enables complex test question generation (e.g. "What is the total monthly rent of all leases \$LESSOR has in \$YEAR and \$MONTH")

\subsubsection{Template-Based Document Generation}
After the ground truth database is created, documents are generated using modular templates with controlled variation. Each template defines the document structure while allowing randomized selection of:
\begin{itemize}
    \item Language style (formal, semi-formal, casual)
    \item Structural organization (section ordering, optional clauses)
    \item Boilerplate text variations
\end{itemize}

This produces documents that are structurally consistent yet superficially diverse, preventing models from exploiting surface-level patterns while maintaining ground truth integrity.

\subsection{Coherent Simulated Universe}

One of the significant shortcomings of synthetic generated data comes from naive generation - that is, when 100 synthetic documents are generated, but they are all \textit{independent generations}. This gives the synthetic data set a characteristic that is very \textit{un-enterprise-like} - the documents are not related to each other, or worse, they have chaotic and unrealistic connections.

For example, in a naive generation, we could generate 100 HR documents using a pool of human names. Being independently randomly generated, our documents (for example, HR evaluations or employee information) would naively randomize facts like employee name, manager, department, etc. The result is a dataset that models no realistic enterprise corpus, for example:
\begin{itemize}
    \item Employees in similar randomized department (by chance) will have a different dept manager or supervisor named
    \item Employees in different departments may accidentally have a similar randomized person name
    \item An employee, named as a manager or supervisor in a particular department from a previous document, can have a different department or position 
\end{itemize}

 The above is not an exhaustive list. This incoherence resulting from naive random generation is a problem. It does not model a realistic enterprise scenario (therefore metrics against that dataset may have very weak correlation to real-world performance), and will prevent the creation of challenging comprehension and aggregation questions, such as \textit{`which manager gave the most evaluations this quarter?'}, because necessary relationships will either not be diverse or coherent enough, or may not even exist at all.

RIKER solves this problem through its \textbf{Coherent Simulated Universe} approach. The very first step in document generation is ground truth creation (see \ref{sec:GroundTruthDB}), and this includes necessary relationships among the different entities. To make this coherence spread across different document types and across the entire universe of generated documents, RIKER has the concept of \textbf{Global Entity Pools}, which are pre-generated and filled as the first step of synthetic data generation. Relationships are then created by drawing from the global entity pools, which are saved in the ground truth database. These global entity pools are used across all document types to create a coherent dataset. For example, a global entity pool called `sales\_agent' contains human names that are used for three types of documents in the current RIKER implementation:
\begin{itemize}
    \item An \textit{optional sales agent} that is named and credited for closing a \textbf{Lease Contract}
    \item A \textit{sales agent} that is named in a \textbf{Sales Agent Field Report}, as the agent who created and submitted the field report detailing sales activities and potential contract status
    \item An \textit{employee} that is named in an \textbf{HR Employee Evaluation} document, as the sales employee being evaluated
\end{itemize}

In RIKER's Coherent Simulated Universe, all the three document types above (in bold) draw from the same Global Entity Pool - meaning the agents you will see who closed lease contracts will be the same agents named in relevant field reports, and named in relevant HR evaluation documents. 

It will also never happen that a Field Report talking about a particular Lease Contract will have a different human name randomized for it as the sales agent - this incoherence is avoided through logic specifically baked-in to the document generation feature, as part of the Coherent Simulated Universe strategy.

This results in having arbitrary scale document generation where ground truth is immediately available with no human annotation effort (because ground truth is where the process actually begins), and the knowledge generated - the entire set of facts and documents - are  coherent according to the generation design.

\subsection{Multi-Level Question Taxonomy}

RIKER generates questions across twelve difficulty levels organized into three categories, each testing distinct capabilities.

\subsubsection{Single-Document Questions (L01 - L04)}

These questions require locating and extracting information from a single document:
\begin{itemize}
    \item \textbf{L01 - Direct Extraction:} Surface-level facts stated explicitly (``What is the monthly rent?'')
    \item \textbf{L02 - Indirect Extraction:} Facts requiring minimal inference (``What is the lease duration?'' when start/end dates are given)
    \item \textbf{L03 - Conditional Extraction:} Facts from optional document sections (``What is the pet deposit?'' — may be N/A)
    \item \textbf{L04 - Complex Extraction:} Facts requiring multiple conditions or cross-referencing within a document
\end{itemize}

\subsubsection{Aggregation Questions (L05 - L10)}

These questions require synthesizing information across multiple documents:
\begin{itemize}
    \item \textbf{L05 - Counting:} ``How many leases does Lessor X have?''
    \item \textbf{L06 - Summation/Averaging:} ``What is the total monthly rent across all leases?"
    \item \textbf{L07 - Comparison:} ``Which lessor has more leases, X or Y?''
    \item \textbf{L08 - Enumeration:} ``List all lessees for Lessor X''
    \item \textbf{L09 - Multi-hop:} ``What is Lessor X's most recent lease end date?''
    \item \textbf{L10 - Temporal:} ``How many leases were active in Q3 2024?''
\end{itemize}

Aggregation questions are particularly challenging because they require the model to: (1) identify all relevant documents, (2) extract the relevant facts from each, and (3) perform the required computation correctly.

\subsubsection{Hallucination Probe Questions (L11 - L12)}

These questions are designed to detect fabrication:
\begin{itemize}
    \item \textbf{L11 - Non-existent Entities:} Questions about entities that do not appear anywhere in the corpus. The entity names are drawn from unused portions of the entity pool, ensuring they are plausible but definitively absent. The only correct response is ``Unknown'' or equivalent.
    \item \textbf{L12 - Absent Information:} Questions about optional fields that are absent from specific documents. For example, asking about the pet deposit for a lease that has no pet clause. The only correct response is ``N/A'' or equivalent.
\end{itemize}

L11 questions are particularly valuable because any specific answer constitutes unambiguous fabrication—the model cannot have retrieved the information from the corpus because it does not exist.

\subsection{Deterministic Scoring}

RIKER employs answer-key-based scoring, eliminating the variability inherent in LLM-as-judge approaches.

\subsubsection{Scoring Mechanisms}

Each question specifies its scoring type:
\begin{itemize}
    \item \textbf{Exact match:} For categorical responses (names, yes/no)
    \item \textbf{Numeric extraction:} Parses numerical answers with tolerance for formatting variations
    \item \textbf{Set comparison:} For enumeration questions, compares answer sets regardless of ordering
    \item \textbf{Semantic equivalence:} For structured responses with known equivalent forms
\end{itemize}

All scoring logic operates against the ground truth database, ensuring reproducibility. The same model outputs will always receive the same scores.

In this particular

\subsubsection{Response Format Enforcement}

Questions include explicit format instructions (e.g., ``Reply with only the number'' or ``Indicate your final answer with: Final answer: [your answer]''). This structured output requirement reduces ambiguity in answer extraction and improves scoring reliability.

\subsection{Fidelity Metrics Taxonomy}

We define a three-level taxonomy for hallucination-related metrics

\subsubsection{Faithfulness (L01 - L04 + L11 - L12)}

The broadest metric, measuring accuracy on all questions where the model had sufficient information to answer correctly. This encompasses both grounding failures (wrong answers from documents that exist) and fabrication (invented information). Faithfulness aligns with the colloquial enterprise definition of ``hallucination'' as any confidently wrong answer when correct information was available.

\subsubsection{Grounding (L01 - L04)}

Accuracy on single-document questions only. Grounding failures indicate the model could not locate or correctly extract information from documents that definitively contain the answer. This isolates retrieval and comprehension errors from fabrication.

\subsubsection{Fabrication (L11 - L12)}

Error rate on hallucination probe questions. Because L11 questions ask about non-existent entities, any specific answer is definitively fabricated—there is no ambiguity about the failure mode. This provides the cleanest signal for measuring a model's tendency to invent information.

\subsubsection{Aggregation (L05 - L10)}

Reported separately as a capability metric rather than a hallucination metric. Aggregation errors conflate multiple failure modes (incomplete document retrieval, computation errors, working memory limitations) that are distinct from hallucination in the traditional sense.

\subsection{Robustness Validation}

A methodology is only useful if it produces stable, reproducible results. We validated RIKER's robustness through cross-corpus experiments.

\subsubsection{Cross-Corpus Stability}

Four corpora were generated from identical configuration parameters but different random seeds, producing documents with different entity names, dates, and surface content while maintaining structural equivalence. Four models spanning different performance tiers were evaluated on all four corpora.

Results demonstrated strong stability: top-performing models showed less than 2\% accuracy variance across corpora (CV $<$ 1\%), with consistent ranking preservation. This validates that RIKER results reflect model capability rather than corpus-specific artifacts.

\subsubsection{Implications for Reproducibility}

The cross-corpus stability finding has practical implications: researchers can generate their own RIKER corpora and expect comparable results to other studies using the same configuration parameters. This addresses a key limitation of static benchmarks: their fixed nature means they inevitably leak into training corpora, while RIKER's regenerability ensures fresh, uncontaminated test data.

\section{Experimental Design}

Table~\ref{tab:experimental-scale} summarizes the scale of our evaluation for this RIKER study. 33 models and over 21B tokens were processed.

\begin{table}[ht]
\centering
\caption{RIKER Experimental Scale}
\label{tab:experimental-scale}
\begin{tabular}{@{}lrrrrrr@{}}
\toprule
\textbf{Corpus} & \textbf{Questions} & \textbf{Runs} & \textbf{Models} & \textbf{Input Tokens} & \textbf{Output Tokens} & \textbf{Total Tokens} \\
\midrule
\multicolumn{7}{@{}l}{\textit{Main Experiment:}} \\
32K   & 110 & 8 & 33 & 0.79B & 7M   & 0.80B \\
128K  & 301 & 8 & 24 & 5.67B & 47M  & 5.72B \\
200K  & 525 & 8 & 11 & 9.26B & 85M  & 9.34B \\
\cmidrule{5-7}
\multicolumn{3}{@{}l}{\textit{Main Subtotal}} & & 15.72B & 139M & 15.86B \\
\midrule
\multicolumn{7}{@{}l}{\textit{Cross-Corpus Validation (Section~\ref{sec:cross-corpus}):}} \\
128K (B,C,D) & 301 & 8 & 4 & 3.02B & 7M & 3.03B \\
\midrule
\multicolumn{7}{@{}l}{\textit{Expanded Hallucination Analysis (Section~\ref{sec:hallucination-analysis}):}} \\
32K (HA,HB,HC,HD) & 300 & 8 & 10 & 2.64B & 9M & 2.65B \\
\midrule
\textbf{Grand Total} & & & & \textbf{21.38B} & \textbf{155M} & \textbf{21.54B} \\
\bottomrule
\end{tabular}

\vspace{0.5em}
\noindent\scriptsize\textit{Token counts reflect total compute consumed across all experimental attempts, including a few runs that failed to produce scorable output due to API errors, timeouts, or malformed responses.}
\end{table}

\begin{table}[ht]
\centering
\caption{Corpus Document Composition}
\label{tab:corpus-composition}
\begin{tabular}{@{}lrrrr@{}}
\toprule
\textbf{Corpus} & \textbf{Leases} & \textbf{Field Reports} & \textbf{HR Reports} & \textbf{Total Docs} \\
\midrule
\multicolumn{5}{@{}l}{\textit{Main Experiment:}} \\
32K   & 10 & 44  & 56  & 110 \\
128K (Set A) & 37 & 216 & 116 & 369 \\
200K  & 60 & 381 & 196 & 637 \\
\midrule
\multicolumn{5}{@{}l}{\textit{Cross-Corpus Validation:}} \\
128K (Set B) & 37 & 255 & 128 & 420 \\
128K (Set C) & 37 & 228 & 120 & 385 \\
128K (Set D) & 37 & 211 & 120 & 368 \\
\bottomrule
\end{tabular}
\end{table}

Table~\ref{tab:corpus-composition} details the document breakdown across corpora; question counts scale proportionally, with approximately 50\% single-document extraction, 40\% cross-document aggregation, and 10\% hallucination probes. All three document types appear in every corpus, demonstrating the Coherent Simulated Universe in practice: the same entities (people, properties, companies) appear across leases, field reports, and HR evaluations. The document distribution reflects realistic business ratios - leases are fewer (one per tenancy), field reports dominate (generated per agent-prospect interaction), and HR reports fall in between (periodic per employee).

Model coverage decreases with context size (33 $\rightarrow$ 24 $\rightarrow$ 11) as fewer models support longer contexts - this stratification is itself a finding. Eight runs per model enable statistical significance testing with variance and confidence interval reporting. 

All experiments use a temperature setting of 0.4, balancing determinism with natural response variation. Future work will explore the effects of LLM temperature on performance in enterprise knowledge extraction settings.

\section{Results}

We evaluate 33 models at 32K context, 24 models at 128K context, and 11 models at 200K context. Results are aggregated across 8 runs per model to enable statistical significance testing.

\subsection{Overall Model Performance}

Tables~\ref{tab:results-32k}, \ref{tab:results-128k}, and \ref{tab:results-200k} present model performance across all three context sizes. Each table reports five metrics:

\begin{itemize}
    \item \textbf{Overall}: Weighted accuracy across all question types.
    \item \textbf{Single-Doc}: Accuracy on questions answerable from a single document (e.g., ``What is the monthly rent in lease X?''). Tests basic retrieval and extraction.
    \item \textbf{Aggregation}: Accuracy on questions requiring synthesis across multiple documents (e.g., ``How many leases does lessor Y have?''). Tests cross-document reasoning.
    \item \textbf{Hall Detect}: Accuracy on hallucination probes - correctly answering ``Unknown'' when asked about non-existent entities. Higher is better.
    \item \textbf{Hall Rate}: Hallucination rate - percentage of probes where the model fabricated an answer instead of admitting uncertainty. \textit{Lower is better.}
\end{itemize}

\begin{table}[H]
\centering
\caption{Model Performance at 32K Context (sorted by overall accuracy)}
\label{tab:results-32k}
\small
\begin{tabular}{@{}lrrrrr@{}}
\toprule
\textbf{Model} & \textbf{Overall} & \textbf{Single-Doc} & \textbf{Aggregation} & \textbf{Hall Detect} & \textbf{Hall Rate} \\
 & \textbf{(\%)}  & \textbf{(\%)} & \textbf{(\%)} & \textbf{(\%)} & \textbf{(\%)} \\
\midrule
GLM-4.5 & 94.7 & 93.6 & 94.6 & 100.0 & 0.0 \\
GLM-4.6 & 90.5 & 93.6 & 86.6 & 89.8 & 10.2 \\
Qwen3-Next-80B & 88.6 & 91.4 & 92.6 & 59.1 & 40.9 \\
Qwen3-235B-FP8 & 87.8 & 86.1 & 95.2 & 67.0 & 33.0 \\
Qwen3-235B & 87.4 & 85.2 & 95.5 & 65.9 & 34.1 \\
Mistral-Large-3 & 86.5 & 90.2 & 85.5 & 71.6 & 28.4 \\
Qwen3-Coder-480B & 86.3 & 84.5 & 86.6 & 93.2 & 6.8 \\
Llama-4-Maverick & 85.5 & 86.6 & 87.5 & 71.6 & 28.4 \\
DeepSeek-V3.1 & 84.9 & 89.5 & 80.4 & 79.5 & 20.5 \\
Qwen2.5-72B & 82.4 & 90.0 & 75.9 & 70.5 & 29.5 \\
Qwen3-30B & 81.1 & 82.3 & 86.4 & 54.5 & 45.5 \\
GLM-4.5-Air & 78.0 & 83.9 & 66.8 & 93.2 & 6.8 \\
Qwen3-Coder-30B & 77.5 & 79.3 & 86.6 & 31.8 & 68.2 \\
Qwen3-4B-Instruct & 77.5 & 86.6 & 69.3 & 64.8 & 35.2 \\
DeepSeek-V3 & 76.1 & 79.5 & 72.4 & 73.9 & 26.1 \\
Llama-3.1-405B & 75.6 & 79.1 & 70.5 & 78.4 & 21.6 \\
Llama-4-Scout & 71.7 & 80.9 & 66.5 & 46.6 & 53.4 \\
Qwen2.5-32B & 71.4 & 83.9 & 53.1 & 81.8 & 18.2 \\
Qwen3-32B & 69.8 & 75.2 & 63.1 & 69.3 & 30.7 \\
Qwen2.5-14B & 67.5 & 75.2 & 61.4 & 53.4 & 46.6 \\
Llama-3.1-70B & 67.4 & 78.0 & 56.0 & 60.2 & 39.8 \\
Qwen2.5-Coder-14B & 63.5 & 77.7 & 51.4 & 40.9 & 59.1 \\
Llama-3.3-70B & 60.9 & 73.4 & 48.9 & 46.6 & 53.4 \\
Qwen3-8B & 60.2 & 75.5 & 38.9 & 69.3 & 30.7 \\
Qwen3-14B & 60.1 & 57.7 & 60.2 & 71.6 & 28.4 \\
Qwen3-4B & 57.7 & 76.1 & 37.8 & 45.5 & 54.5 \\
Qwen2.5-Coder-7B & 53.4 & 62.0 & 40.3 & 62.5 & 37.5 \\
Granite-4-Small & 49.7 & 63.2 & 34.9 & 40.9 & 59.1 \\
Llama-3.1-8B & 48.5 & 64.5 & 31.3 & 37.5 & 62.5 \\
Granite-4-Tiny & 38.2 & 49.5 & 31.8 & 6.8 & 93.2 \\
Llama-3.2-3B & 30.3 & 36.8 & 14.2 & 62.5 & 37.5 \\
Granite-4-Micro & 26.9 & 38.9 & 14.2 & 18.2 & 81.8 \\
Llama-3.2-1B & 9.1 & 4.5 & 3.4 & 54.5 & 45.5 \\
\bottomrule
\end{tabular}
\end{table}

\begin{table}[H]
\centering
\caption{Model Performance at 128K Context (sorted by overall accuracy)}
\label{tab:results-128k}
\small
\begin{tabular}{@{}lrrrrr@{}}
\toprule
\textbf{Model} & \textbf{Overall} & \textbf{Single-Doc} & \textbf{Aggregation} & \textbf{Hall Detect} & \textbf{Hall Rate} \\
 & \textbf{(\%)}  & \textbf{(\%)} & \textbf{(\%)} & \textbf{(\%)} & \textbf{(\%)} \\
\midrule
Qwen3-Next-80B & 88.5 & 94.0 & 80.8 & 90.4 & 9.6 \\
Qwen3-235B & 84.6 & 92.3 & 75.5 & 80.0 & 20.0 \\
Qwen3-235B-FP8 & 84.6 & 92.1 & 76.1 & 78.3 & 21.7 \\
GLM-4.5 & 84.4 & 94.8 & 68.5 & 92.1 & 7.9 \\
DeepSeek-V3.1 & 84.1 & 94.0 & 72.0 & 79.6 & 20.4 \\
GLM-4.6 & 81.4 & 91.4 & 69.1 & 77.1 & 22.9 \\
Qwen3-Coder-480B & 80.7 & 87.6 & 70.6 & 84.2 & 15.8 \\
Mistral-Large-3 & 75.6 & 89.4 & 61.3 & 60.0 & 40.0 \\
Qwen3-30B & 71.7 & 82.7 & 61.9 & 52.9 & 47.1 \\
DeepSeek-V3 & 70.2 & 88.3 & 46.2 & 69.6 & 30.4 \\
Llama-4-Maverick & 68.2 & 83.4 & 51.9 & 52.9 & 47.1 \\
GLM-4.5-Air & 67.3 & 84.4 & 40.5 & 82.1 & 17.9 \\
Qwen3-Coder-30B & 59.9 & 71.7 & 49.8 & 38.3 & 61.7 \\
Qwen3-4B & 57.7 & 75.0 & 32.0 & 67.5 & 32.5 \\
Llama-3.1-405B & 52.7 & 69.9 & 26.6 & 64.6 & 35.4 \\
Llama-4-Scout & 51.5 & 66.5 & 30.6 & 54.6 & 45.4 \\
Llama-3.1-70B & 45.1 & 65.2 & 17.9 & 46.7 & 53.3 \\
Llama-3.3-70B & 40.8 & 55.2 & 18.6 & 52.1 & 47.9 \\
Granite-4-Small & 39.9 & 54.0 & 15.7 & 60.4 & 39.6 \\
Llama-3.1-8B & 35.3 & 51.1 & 16.0 & 28.8 & 71.3 \\
Llama-3.2-3B & 25.9 & 33.1 & 7.1 & 61.3 & 38.8 \\
Granite-4-Tiny & 22.6 & 32.1 & 14.9 & 3.8 & 96.3 \\
Granite-4-Micro & 22.5 & 25.7 & 9.8 & 55.0 & 45.0 \\
Llama-3.2-1B & 11.3 & 7.8 & 1.6 & 67.1 & 32.9 \\
\bottomrule
\end{tabular}
\end{table}

\begin{table}[H]
\centering
\caption{Model Performance at 200K Context (sorted by overall accuracy)}
\label{tab:results-200k}
\small
\begin{tabular}{@{}lrrrrr@{}}
\toprule
\textbf{Model} & \textbf{Overall} & \textbf{Single-Doc} & \textbf{Aggregation} & \textbf{Hall Detect} & \textbf{Hall Rate} \\
 & \textbf{(\%)}  & \textbf{(\%)} & \textbf{(\%)} & \textbf{(\%)} & \textbf{(\%)} \\
\midrule
Qwen3-Next-80B & 77.6 & 88.8 & 59.6 & 84.8 & 15.2 \\
Qwen3-235B & 72.3 & 83.5 & 55.5 & 75.5 & 24.5 \\
Qwen3-235B-FP8 & 71.6 & 82.1 & 56.0 & 74.5 & 25.5 \\
Qwen3-Coder-480B & 71.1 & 81.4 & 55.2 & 75.9 & 24.1 \\
Mistral-Large-3 & 68.3 & 80.5 & 52.4 & 63.6 & 36.4 \\
Qwen3-30B & 65.3 & 75.1 & 54.1 & 56.4 & 43.6 \\
Llama-4-Maverick & 62.3 & 75.5 & 46.9 & 51.1 & 48.9 \\
Qwen3-Coder-30B & 52.8 & 67.9 & 35.6 & 38.2 & 61.8 \\
Llama-4-Scout & 47.6 & 62.5 & 26.5 & 47.7 & 52.3 \\
Qwen3-4B & 42.1 & 54.3 & 17.8 & 67.0 & 33.0 \\
GLM-4.6 & 34.3 & 47.3 & 19.6 & 21.4 & 78.6 \\
\bottomrule
\end{tabular}
\end{table}

\subsection{Performance by Question Category}

Performance varies dramatically by question type. Single-document extraction (finding a specific fact in one document) is consistently the easiest task, with top models exceeding 90\%. Aggregation queries (counting, comparing across documents) prove significantly harder, with even the best model achieving only 80.8\%. Hallucination detection (correctly rejecting queries about non-existent entities) reveals the starkest differences between models.

Key observations:
\begin{itemize}
    \item Single-document extraction is consistently highest - pattern-matching often suffices
    \item Aggregation queries are significantly harder - requires reasoning across multiple documents
    \item Hallucination rates vary from 7.9\% (GLM-4.5) to 96.3\% (Granite-4-Tiny)
    \item Some models (Granite-4-Tiny) hallucinate on nearly every probe question
\end{itemize}

\subsection{Scaling Effects}

Table~\ref{tab:scaling} shows performance degradation as context length increases for models tested at multiple context sizes.

\begin{table}[ht]
\centering
\caption{Performance Degradation Across Context Sizes (models tested at 3 sizes)}
\label{tab:scaling}
\small
\begin{tabular}{@{}lrrrl@{}}
\toprule
\textbf{Model} & \textbf{32K (\%)} & \textbf{128K (\%)} & \textbf{200K (\%)} & \textbf{$\Delta$ (32K$\rightarrow$200K)} \\
\midrule
Qwen3-Next-80B & 88.6 & 88.5 & 77.6 & $-$11.0 \\
Qwen3-235B & 87.4 & 84.6 & 72.3 & $-$15.1 \\
Qwen3-Coder-480B & 86.3 & 80.7 & 71.1 & $-$15.2 \\
Qwen3-30B-A3B-Instruct & 81.1 & 71.7 & 65.3 & $-$15.8 \\
Qwen3-235B-FP8 & 87.8 & 84.6 & 71.6 & $-$16.2 \\
Mistral-Large-3 & 86.5 & 75.6 & 68.3 & $-$18.2 \\
Llama-4-Maverick & 85.5 & 68.2 & 62.3 & $-$23.2 \\
Qwen3-Coder-30B & 77.5 & 59.9 & 52.8 & $-$24.7 \\
Llama-4-Scout & 71.7 & 51.5 & 47.6 & $-$24.1 \\
Qwen3-4B & 77.5 & 57.7 & 42.1 & $-$35.4 \\
GLM-4.6 & 90.5 & 81.4 & 34.3 & $-$56.2 \\
\bottomrule
\end{tabular}
\end{table}

Table~\ref{tab:scaling-partial} shows models tested at 32K and 128K only.

\begin{table}[ht]
\centering
\caption{Performance Degradation: 32K to 128K (models not tested at 200K)}
\label{tab:scaling-partial}
\small
\begin{tabular}{@{}lrrl@{}}
\toprule
\textbf{Model} & \textbf{32K (\%)} & \textbf{128K (\%)} & \textbf{$\Delta$} \\
\midrule
GLM-4.5 & 94.7 & 84.4 & $-$10.3 \\
DeepSeek-V3.1 & 84.9 & 84.1 & $-$0.8 \\
GLM-4.5-Air & 78.0 & 67.3 & $-$10.7 \\
DeepSeek-V3 & 76.1 & 70.2 & $-$5.9 \\
Llama-3.1-405B & 75.6 & 52.7 & $-$22.9 \\
Llama-3.1-70B & 67.4 & 45.1 & $-$22.3 \\
Llama-3.3-70B & 60.9 & 40.8 & $-$20.1 \\
Granite-4-Small & 49.7 & 39.9 & $-$9.8 \\
Llama-3.1-8B & 48.5 & 35.3 & $-$13.2 \\
Granite-4-Tiny & 38.2 & 22.6 & $-$15.6 \\
Llama-3.2-3B & 30.3 & 25.9 & $-$4.4 \\
Granite-4-Micro & 26.9 & 22.5 & $-$4.4 \\
Llama-3.2-1B & 9.1 & 11.3 & $+$2.2 \\
\bottomrule
\end{tabular}
\end{table}

Notable findings:
\begin{itemize}
    \item GLM-4.6 shows dramatic collapse at 200K ($-$56.2\% from 32K)
    \item Llama models degrade 13 - 23\% from 32K to 128K - consistently worse than other families
    \item Top-tier models (Qwen3, GLM-4.5 family) lose 10-16\% accuracy from 32K to 200K, while others degrade more substantially
    \item Llama-3.2-1B anomalously \textit{improves} at 128K ($+$2.2\%), though from a very low baseline (9.1\%), most likely because its performance is no different than random guessing even from the low-context 32K scenario.
\end{itemize}

\subsection{Model Family Patterns}

\textbf{Qwen3 Family:} Dominates the leaderboard. Qwen3-Next-80B achieves the best overall performance (88.5\% at 128K). The 235B variants show strong performance but the smaller 30B and 4B models remain competitive for their size.

\textbf{DeepSeek:} V3.1 significantly outperforms V3 (84.1\% vs 70.2\%), suggesting meaningful architectural or training improvements between versions.

\textbf{GLM:} GLM-4.5 excels at hallucination detection (92.1\% detection rate, only 7.9\% hallucination rate) but GLM-4.6 shows catastrophic failure at 200K context.

\textbf{Llama:} The Llama 3.x family underperforms relative to model size. Llama-3.1-405B (52.7\%) is outperformed by much smaller Qwen models. Llama 4 marks a significant improvement - Maverick (85.5\% at 32K) outperforms Llama-3.1-405B by 10 points, but still gets outperformed by much smaller Qwen models at 128K and higher.

\textbf{Granite:} IBM's Granite models struggle significantly, with the tiny variant showing 96.3\% hallucination rate - fabricating answers to nearly every hallucination probe.

\textbf{Mistral}: Mistral-Large-3 performs competitively (86.5\% at 32K, 68.3\% at 200K) but shows steeper degradation at longer contexts than Qwen models.

\subsection{Hallucination Analysis}

We decompose model hallucination measures into three nested metrics:

\begin{itemize}
    \item \textbf{Faithfulness} (L01 - L04 + L11 - L12): Any error where the model had sufficient information to answer correctly. Combines single-document extraction and hallucination probe questions. Higher is better.
    \item \textbf{Grounding} (L01 - L04): Single-document questions only - measures whether the model can find and correctly read the relevant document. Higher is better.
    \item \textbf{Fabrication} (L11 - L12): Hallucination probe questions using non-existent entities. Any specific answer is \textit{definitively} fabricated since the entities exist nowhere in the corpus. Lower is better.
\end{itemize}

Aggregation questions (L05 - L10) are excluded from hallucination metrics because errors conflate grounding failures with computation errors and incoherence loss - these are synthesis failures, not hallucination in any useful sense. Aggregation performance is analyzed separately in Section~\ref{sec:aggregation}.

Tables~\ref{tab:hallucination-32k}, \ref{tab:hallucination-128k}, and \ref{tab:hallucination-200k} present the hallucination metrics across all three context sizes.

\begin{table}[ht]
\centering
\begin{minipage}[t]{0.48\textwidth}
\centering
\caption{Hallucination Metrics at 32K Context}
\label{tab:hallucination-32k}
\scriptsize
\begin{tabular}{@{}lrrr@{}}
\toprule
\textbf{Model} & \textbf{Faith.} & \textbf{Ground.} & \textbf{Fab.} \\
 & \textbf{(\%)} & \textbf{(\%)} & \textbf{(\%)} \\
\midrule
GLM-4.5 & 94.7 & 93.6 & 0.0 \\
GLM-4.6 & 93.0 & 93.6 & 10.2 \\
DeepSeek-V3.1 & 87.9 & 89.5 & 20.5 \\
Mistral-Large-3 & 87.1 & 90.2 & 28.4 \\
Qwen2.5-72B & 86.7 & 90.0 & 29.5 \\
Qwen3-Next-80B & 86.0 & 91.4 & 40.9 \\
Qwen3-Coder-480B & 86.0 & 84.5 & 6.8 \\
GLM-4.5-Air & 85.4 & 83.9 & 6.8 \\
Llama-4-Maverick & 84.1 & 86.6 & 28.4 \\
Qwen2.5-32B & 83.5 & 83.9 & 18.2 \\
Qwen3-4B-Instruct & 83.0 & 86.6 & 35.2 \\
Qwen3-235B-FP8 & 83.0 & 86.1 & 33.0 \\
Qwen3-235B & 82.0 & 85.2 & 34.1 \\
Llama-3.1-405B & 79.0 & 79.1 & 21.6 \\
DeepSeek-V3 & 78.6 & 79.5 & 26.1 \\
Qwen3-30B-A3B-Instruct & 77.7 & 82.3 & 45.5 \\
Llama-4-Scout & 75.2 & 80.9 & 53.4 \\
Llama-3.1-70B & 75.0 & 78.0 & 39.8 \\
Qwen3-8B & 74.4 & 75.5 & 30.7 \\
Qwen3-32B & 74.2 & 75.2 & 30.7 \\
Qwen2.5-Coder-14B & 71.6 & 77.7 & 59.1 \\
Qwen2.5-14B & 71.6 & 75.2 & 46.6 \\
Qwen3-Coder-30B & 71.4 & 79.3 & 68.2 \\
Qwen3-4B & 71.0 & 76.1 & 54.5 \\
Llama-3.3-70B & 68.9 & 73.4 & 53.4 \\
Qwen2.5-Coder-7B & 62.1 & 62.0 & 37.5 \\
Llama-3.1-8B & 60.0 & 64.5 & 62.5 \\
Qwen3-14B & 60.0 & 57.7 & 28.4 \\
Granite-4-Small & 59.5 & 63.2 & 59.1 \\
Granite-4-Tiny & 42.4 & 49.5 & 93.2 \\
Llama-3.2-3B & 41.1 & 36.8 & 37.5 \\
Granite-4-Micro & 35.4 & 38.9 & 81.8 \\
Llama-3.2-1B & 12.9 & 4.5 & 45.5 \\
\bottomrule
\end{tabular}
\end{minipage}%
\hfill
\begin{minipage}[t]{0.50\textwidth}
\centering
\caption{Hallucination Metrics at 128K Context}
\label{tab:hallucination-128k}
\scriptsize
\begin{tabular}{@{}lrrr@{}}
\toprule
\textbf{Model} & \textbf{Faith.} & \textbf{Ground.} & \textbf{Fab.} \\
 & \textbf{(\%)} & \textbf{(\%)} & \textbf{(\%)} \\
\midrule
GLM-4.5 & 94.4 & 94.8 & 7.9 \\
Qwen3-Next-80B & 93.4 & 94.0 & 9.6 \\
DeepSeek-V3.1 & 91.6 & 94.0 & 20.4 \\
Qwen3-235B & 90.3 & 92.3 & 20.0 \\
Qwen3-235B-FP8 & 89.9 & 92.1 & 21.7 \\
GLM-4.6 & 89.1 & 91.4 & 22.9 \\
Qwen3-Coder-480B & 87.0 & 87.6 & 15.8 \\
DeepSeek-V3 & 85.3 & 88.3 & 30.4 \\
Mistral-Large-3 & 84.6 & 89.4 & 40.0 \\
GLM-4.5-Air & 84.1 & 84.4 & 17.9 \\
Llama-4-Maverick & 78.4 & 83.4 & 47.1 \\
Qwen3-30B-A3B-Instruct & 77.9 & 82.7 & 47.1 \\
Qwen3-4B & 73.8 & 75.0 & 32.5 \\
Llama-3.1-405B & 69.1 & 69.9 & 35.4 \\
Qwen3-Coder-30B & 66.3 & 71.7 & 61.7 \\
Llama-4-Scout & 64.5 & 66.5 & 45.4 \\
Llama-3.1-70B & 62.2 & 65.2 & 53.3 \\
Granite-4-Small & 55.1 & 54.0 & 39.6 \\
Llama-3.3-70B & 54.7 & 55.2 & 47.9 \\
Llama-3.1-8B & 47.5 & 51.1 & 71.3 \\
Llama-3.2-3B & 37.7 & 33.1 & 38.8 \\
Granite-4-Micro & 30.5 & 25.7 & 45.0 \\
Granite-4-Tiny & 27.5 & 32.1 & 96.3 \\
Llama-3.2-1B & 17.4 & 7.8 & 32.9 \\
\bottomrule
\end{tabular}

\vspace{1em}

\caption{Hallucination Metrics at 200K Context}
\label{tab:hallucination-200k}
\scriptsize
\begin{tabular}{@{}lrrr@{}}
\toprule
\textbf{Model} & \textbf{Faith.} & \textbf{Ground.} & \textbf{Fab.} \\
 & \textbf{(\%)} & \textbf{(\%)} & \textbf{(\%)} \\
\midrule
Qwen3-Next-80B & 88.1 & 88.8 & 15.2 \\
Qwen3-235B & 82.2 & 83.5 & 24.5 \\
Qwen3-235B-FP8 & 80.9 & 82.1 & 25.5 \\
Qwen3-Coder-480B & 80.5 & 81.4 & 24.1 \\
Mistral-Large-3 & 77.7 & 80.5 & 36.4 \\
Qwen3-30B-A3B-Instruct & 72.0 & 75.1 & 43.6 \\
Llama-4-Maverick & 71.4 & 75.5 & 48.9 \\
Qwen3-Coder-30B & 63.0 & 67.9 & 61.8 \\
Llama-4-Scout & 60.0 & 62.5 & 52.3 \\
Qwen3-4B & 56.4 & 54.3 & 33.0 \\
GLM-4.6 & 43.0 & 47.3 & 78.6 \\
\bottomrule
\end{tabular}
\end{minipage}
\end{table}

\textbf{Key findings:}

\begin{itemize}
    \item \textbf{GLM-4.5 dominates at shorter contexts:} At 32K, GLM-4.5 achieves 94.7\% faithfulness with \textit{zero} fabrication - perfect hallucination resistance. This degrades to 7.9\% fabrication at 128K, still best-in-class. GLM-4.5 is not tested at 200K due to its native max context limitation.

    \item \textbf{Fabrication rates increase with context:} Most models show higher fabrication at longer contexts. GLM-4.6 demonstrates this dramatically: 10.2\% fabrication at 32K rises to 78.6\% at 200K - a 7.7$\times$ increase that explains its overall collapse.

    \item \textbf{Qwen3-Next-80B dominates long context:} Fabrication starts very high, but vastly decreases in higher context sizes (40.9\% $\rightarrow$ 9.6\% $\rightarrow$ 15.2\% for 32K/128K/200K). The 32K anomaly is most likely a test measurement limitation, as the test set for the 32K corpus has fewer  questions, and thus even fewer hallucination-specific questions as composition. In the 200K context test, Qwen3-Next-80B dominates, and is second only to GLM-4.5 in 128K.

    \item \textbf{Grounding gaps reveal fabrication:} The difference between Faithfulness and Grounding quantifies fabrication's impact. Models with high grounding but high fabrication (e.g., DeepSeek-V3 at 128K: 88.3\% grounding, 30.4\% fabrication) can read documents correctly but invent information when asked about non-existent entities.
\end{itemize}

\textbf{Enterprise implications:} High fabrication rates indicate models that confidently invent information when asked about non-existent entities. In production, such fabrications are indistinguishable from correct answers without ground truth verification. The 96.3\% fabrication rate of Granite-4-Tiny at 128K means it would fabricate answers to nearly every query about entities not in its context - a severe reliability risk. More concerning is the fabrication increase with context length: GLM-4.6's jump from 10.2\% to 78.6\% fabrication suggests that models reliable at shorter contexts can become extremely unreliable at longer ones. Exhaustive testing is warranted.

\subsection{Aggregation Analysis}\label{sec:aggregation}

Aggregation queries require models to synthesize information across multiple documents - counting entities, comparing values, or computing statistics. Unlike single-document extraction (pattern matching) or hallucination probes (fabrication detection), aggregation tests cross-document reasoning.

Table~\ref{tab:aggregation} presents aggregation accuracy across all three context sizes, sorted by 32K performance. Table~\ref{tab:truncation} shows coherence loss rates for aggregation queries (see Section~\ref{sec:coherence-loss} for coherence loss discussion).

\begin{table}[ht]
\centering
\begin{minipage}[t]{0.48\textwidth}
\centering
\captionof{table}{Aggregation Accuracy (sorted by 32K)\label{tab:aggregation}}
\small
\begin{tabular}{@{}lrrr@{}}
\toprule
\textbf{Model} & \textbf{32K} & \textbf{128K} & \textbf{200K} \\
\midrule
Qwen3-235B & 95.5 & 75.5 & 55.5 \\
Qwen3-235B-FP8 & 95.2 & 76.1 & 56.0 \\
GLM-4.5 & 94.6 & 68.5 & - \\
Qwen3-Next-80B & 92.6 & 80.8 & 59.6 \\
Llama-4-Maverick & 87.5 & 51.9 & 46.9 \\
GLM-4.6 & 86.7 & 69.1 & 19.6 \\
Qwen3-Coder-480B & 86.7 & 70.6 & 55.2 \\
Qwen3-Coder-30B & 86.7 & 49.8 & 35.6 \\
Qwen3-30B-A3B-Instruct & 86.4 & 61.9 & 54.1 \\
Mistral-Large-3 & 85.5 & 61.3 & 52.4 \\
DeepSeek-V3.1 & 80.4 & 72.0 & - \\
Qwen2.5-72B & 75.9 & - & - \\
DeepSeek-V3 & 72.4 & 46.2 & - \\
Llama-3.1-405B & 70.5 & 26.6 & - \\
Qwen3-4B-Instruct & 69.3 & 57.7 & 17.8 \\
GLM-4.5-Air & 66.8 & 40.5 & - \\
Llama-4-Scout & 66.5 & 30.6 & 26.5 \\
Qwen3-32B & 63.1 & - & - \\
Qwen2.5-14B & 61.4 & - & - \\
Llama-3.1-70B & 56.0 & 17.9 & - \\
Qwen2.5-32B & 53.1 & - & - \\
Llama-3.3-70B & 48.9 & 18.6 & - \\
Qwen3-8B & 38.9 & - & - \\
Qwen3-4B & 37.8 & - & - \\
Llama-3.1-8B & 31.3 & 16.0 & - \\
Granite-4-Small & 34.9 & 15.7 & - \\
Granite-4-Tiny & 31.8 & 14.9 & - \\
Granite-4-Micro & 14.2 & 9.8 & - \\
Llama-3.2-3B & 14.2 & 7.1 & - \\
Llama-3.2-1B & 3.4 & 1.6 & - \\
\bottomrule
\end{tabular}

\vspace{0.3em}
\noindent\scriptsize\textit{Values in \%. Dash (-) = not tested.}
\end{minipage}%
\hfill
\begin{minipage}[t]{0.48\textwidth}
\centering
\captionof{table}{Coherence Loss\label{tab:truncation}}
\small
\begin{tabular}{@{}lrrr@{}}
\toprule
\textbf{Model} & \textbf{32K} & \textbf{128K} & \textbf{200K} \\
\midrule
\multicolumn{4}{@{}l}{\textit{Severe ($>$10\%):}} \\
Qwen3-4B-Instruct & 0.3 & 11.3 & 37.0 \\
Qwen3-30B-A3B-Instruct & 0.0 & 2.8 & 15.9 \\
Llama-3.1-8B & 0.3 & 13.9 & - \\
Qwen3-Coder-30B & 0.0 & 1.5 & 13.4 \\
Qwen3-Next-80B & 0.0 & 0.3 & 13.3 \\
GLM-4.6 & 0.0 & 0.4 & 11.6 \\
\midrule
\multicolumn{4}{@{}l}{\textit{Moderate (5 - 10\%):}} \\
Llama-3.2-1B & 9.6 & 6.0 & - \\
Llama-3.2-3B & 0.0 & 7.8 & - \\
Granite-4-Tiny & 3.4 & 7.7 & - \\
Llama-3.3-70B & 0.0 & 5.4 & - \\
Qwen3-4B & 5.1 & - & - \\
\midrule
\multicolumn{4}{@{}l}{\textit{Minor (1 - 5\%):}} \\
Granite-4-Micro & 0.0 & 4.4 & - \\
Qwen3-235B-FP8 & 0.0 & 0.5 & 3.8 \\
Qwen3-235B & 0.0 & 0.9 & 3.5 \\
GLM-4.5-Air & 0.0 & 1.8 & - \\
Granite-4-Small & 0.3 & 1.6 & - \\
Qwen3-8B & 1.4 & - & - \\
\midrule
\multicolumn{4}{@{}l}{\textit{Rare ($<$1\%):}} \\
Llama-3.1-70B & 0.0 & 0.9 & - \\
Llama-3.1-405B & 0.6 & 0.4 & - \\
Qwen3-14B & 0.6 & - & - \\
Qwen3-32B & 0.6 & - & - \\
DeepSeek-V3 & 0.0 & 0.1 & - \\
DeepSeek-V3.1 & 0.0 & 0.0 & - \\
GLM-4.5 & 0.0 & 0.0 & - \\
Llama-4-Scout & 0.0 & 0.0 & 0.5 \\
Qwen3-Coder-480B & 0.0 & 0.0 & 0.0 \\
Mistral-Large-3 & 0.0 & 0.0 & 0.0 \\
Llama-4-Maverick & 0.0 & 0.0 & 0.0 \\

Qwen2.5-72B & 0.0 & - & - \\
Qwen2.5-32B & 0.0 & - & - \\
Qwen2.5-14B & 0.0 & - & - \\
Qwen2.5-Coder-14B & 0.0 & - & - \\
Qwen2.5-Coder-7B & 0.0 & - & - \\
\bottomrule
\end{tabular}

\vspace{0.3em}
\noindent\scriptsize\textit{Values in \%. Dash (-) = not tested.}
\end{minipage}
\end{table}

\textbf{Key findings:}

\begin{itemize}
    \item \textbf{Aggregation is consistently harder than single-document extraction:} Even the best model (Qwen3-235B) drops from 92.3\% grounding to 75.5\% aggregation at 128K - a 17-point gap. Cross-document reasoning is fundamentally more difficult.

    \item \textbf{Aggregation degrades faster with context:} Qwen3-Next-80B drops from 92.6\% (32K) to 80.8\% (128K) to 59.6\% (200K) - a 33-point decline. Compare this to its faithfulness decline of only 5 points across the same range.

    \item \textbf{Llama family struggles disproportionately:} Llama-3.1-405B achieves 79.0\% faithfulness at 32K but only 70.5\% aggregation. At 128K, this gap widens: 69.1\% faithfulness vs 26.6\% aggregation - a 42-point disparity.

    \item \textbf{GLM-4.6 collapse is aggregation-specific:} GLM-4.6 drops from 86.7\% aggregation at 32K to 19.6\% at 200K (67-point drop), while its faithfulness drops from 93.0\% to 43.0\% (50-point drop). The aggregation failure is even more severe.

    \item \textbf{Some models maintain aggregation better:} Qwen3-Coder-480B shows remarkable stability: 86.7\% $\rightarrow$ 70.6\% $\rightarrow$ 55.2\% across 32K/128K/200K - consistent 15-16 point drops per tier, compared to the erratic collapses seen in other models.
\end{itemize}

\textbf{Enterprise implications:} Aggregation queries are common in enterprise knowledge systems: ``How many contracts expire this quarter?'', ``What is the total revenue across all subsidiaries?'', ``Which employees have pending reviews?'' The steep performance degradation at longer contexts suggests that production systems should either (1) keep context windows conservative for aggregation tasks, (2) use retrieval strategies that minimize context size, or (3) explicitly decompose aggregation queries into multiple single-document lookups.

\subsection{Coherence Loss and Infinite Generation}\label{sec:coherence-loss}

Under test conditions, our examined models have anywhere from 25K to $>$100K max output tokens available (depending on the model's capacity and the test set). When models hit their maximum token limit before completing a response, this typically indicates an \textit{infinite generation loop} caused by coherence loss. Our benchmarking infrastructure truncates the response, records it, and moves on to the next item. We track truncation rates as a proxy for coherence loss under long-context pressure. Table~\ref{tab:truncation} shows coherence loss organized by severity tier.

\textbf{Key observations from Table~\ref{tab:truncation}:}

\begin{itemize}
    \item \textbf{Coherence loss correlates with model size:} Smaller models (Qwen3-4B, Llama-3.1-8B, Granite-4-Tiny) show higher coherence loss rates. The largest models (Qwen3-Coder-480B, Qwen3-235B, DeepSeek-V3.1) rarely devolve into incoherence.

    \item \textbf{Coherence loss increases with context length:} Qwen3-4B-Instruct goes from 0.3\% at 32K to 37.0\% at 200K - a 123× increase.

    \item \textbf{Aggregation is uniquely vulnerable:} Single-document and hallucination queries rarely trigger coherence loss ($<$0.5\% for most models). Aggregation queries require the model to enumerate and reason across multiple documents, which appears to trigger generation loops.

    \item \textbf{GLM-4.6's 200K collapse has multiple causes:} Beyond the 78.6\% fabrication rate noted earlier, GLM-4.6 also shows 11.6\% coherence loss frequency at 200K - the model fails in multiple ways simultaneously.

\end{itemize}

\textbf{Note: Hardware-dependent failures observed distinct from coherence loss:} Llama 4 models (Maverick and Scout) show near-zero truncation rates, but exhibited a different failure mode on MI300X hardware (producing only ASCII replacement characters in an infinite generation loop) and Gaudi 3 (real characters, but 100\% coherence loss). Successful runs were only possible from our H200 hardware. This infrastructure sensitivity represents a separate reliability dimension not captured by coherence metrics. We note this here for completeness. When running properly, Llama 4 models appear less prone to infinite generation. When they don't, they appear completely broken under RIKER tests - a contradiction the coherence loss metrics in Table~\ref{tab:truncation} cannot capture. None of the other 31 models tested had similar characteristics.

\subsection{Examining the GLM 4.6 200K Collapse}

By all measures - accuracy, hallucination metrics, coherence loss - GLM 4.6 showed a steep decline in output quality in the 200K tests. We investigated whether an adjustment in temperature would change this catastrophic performance, particularly the coherence loss.

\begin{table}[ht]
\centering
\caption{GLM-4.6 at 200K: Temperature 0.4 vs 1.0}
\label{tab:temperature-ablation}
\small
\begin{tabular}{@{}lrr@{}}
\toprule
\textbf{Metric} & \textbf{temp=0.4} & \textbf{temp=1.0} \\
\midrule
Coherence Loss & 184 & 4 \\
Coherence Loss Rate & 4.38\% & 0.10\% \\
Overall Accuracy & 37\% & 27\% \\
\bottomrule
\end{tabular}
\end{table}

Higher temperature dramatically reduces coherence loss (184 $\rightarrow$ 4 instances). Surprisingly, accuracy \textit{decreased} from 37\% to 27\%.

The role of temperature in enterprise-relevant scenarios like long-context knowledge extraction (this study) or in overall agentic fitness (measured in our agentic merit index paper \cite{roig2025standardenterpriserelevantagenticai} with deeper analysis in our trace-level agentic AI failure study \cite{roig2025llmsfailagenticscenarios}) deserves a more intensive treatment and will be examined in much deeper detail in future work.

\subsection{Cross-Corpus Benchmark Stability Analysis}\label{sec:cross-corpus}

A key claim of RIKER is that the Coherent Simulated Universe approach measures \textit{model capability} rather than corpus-specific artifacts. To validate this, we generated three additional 128K corpora (Sets B, C, D) using identical generation parameters to the original (Set A) - same entity counts, document distributions, and question type ratios - but with independent random seeds producing entirely different documents, entities, and questions.

We ran four models spanning the performance spectrum across all three new corpora: DeepSeek-V3.1 and GLM-4.5 (top tier), Qwen3-Coder-480B (upper-middle tier), and Granite-4-Small (lower tier). Table~\ref{tab:cross-corpus} presents the results.

\begin{table}[ht]
\centering
\caption{Cross-Corpus Stability: Model Performance Across Independent 128K Corpora}
\label{tab:cross-corpus}
\small
\begin{tabular}{@{}lrrrrrr@{}}
\toprule
\textbf{Model} & \textbf{Set A} & \textbf{Set B} & \textbf{Set C} & \textbf{Set D} & \textbf{Mean} & \textbf{Spread} \\
 & \textbf{(\%)} & \textbf{(\%)} & \textbf{(\%)} & \textbf{(\%)} & \textbf{(\%)} & \textbf{(pts)} \\
\midrule
DeepSeek-V3.1 & 84.1 & 83.9 & 84.4 & 82.9 & 83.8 & 1.5 \\
GLM-4.5 & 84.4 & 83.0 & 83.6 & 84.4 & 83.8 & 1.5 \\
Qwen3-Coder-480B & 80.7 & 78.9 & 79.9 & 81.4 & 80.2 & 2.5 \\
Granite-4-Small & 39.9 & 36.3 & 33.8 & 34.9 & 36.2 & 6.1 \\
\bottomrule
\end{tabular}

\noindent\small\textit{Note: Set A is the original corpus used throughout this paper. Sets B, C, D are independently generated validation corpora. Spread is max minus min across all tested sets.}
\end{table}

\textbf{Key findings:}

\begin{itemize}
    \item \textbf{Top-tier models show remarkable stability:} DeepSeek-V3.1 and GLM-4.5 vary by only 1.4-1.5 percentage points across four completely independent corpora. This spread is comparable to run-to-run variance within a single corpus.

    \item \textbf{Mid-tier stability is also strong:} Qwen3-Coder-480B shows 2.5 points of spread - slightly higher but still within normal measurement noise for 8-run experiments.

    \item \textbf{Weaker models show higher corpus sensitivity:} Granite-4-Small exhibits 6.1 points of spread. However, its relative ranking is preserved - it remains clearly in the lower performance tier across all corpora. This suggests that weaker models may be more sensitive to specific document phrasings or entity configurations.

    \item \textbf{Ranking stability matters most:} The practical question for benchmarks is whether relative model rankings are preserved. Across all four corpora, the ordering DeepSeek-V3.1 $\approx$ GLM-4.5 > Qwen3-Coder-480B >> Granite-4-Small remains consistent.
\end{itemize}

These results show a simple validation of RIKER's core design. Because ground truth is generated \textit{before} documents, and documents are generated \textit{from} that ground truth, any corpus instantiation with the same parameters measures the same underlying capability. The benchmark is not testing whether models memorized specific phrasings or entity names - it is testing whether they can extract structured information from realistic enterprise documents. This property enables contamination-resistant evaluation: even if a specific corpus leaks, regenerating with the same parameters produces an equally valid benchmark.

\subsection{Expanded Hallucination Analysis}\label{sec:hallucination-analysis}

Our main experiment includes hallucination questions (L11-L12) as part of the broader question taxonomy. To enable deeper analysis of hallucination behavior, we conducted an expanded study with 300 questions per evaluation: 150 grounding questions (L1-L4) testing extraction of facts that \textit{do} exist in the corpus, and 150 fabrication questions (L11-L12) testing whether models correctly identify facts that do \textit{not} exist.

We evaluated 10 models across four independent question sets (HA, HB, HC, HD) on the same 128K corpus, with 8 runs per configuration. This design serves two purposes: (1) deeper hallucination analysis with balanced grounding/fabrication questions, and (2) within-corpus stability validation using different question sets on identical documents.

\begin{table}[ht]
\centering
\caption{Expanded Hallucination Analysis: Grounding vs.\ Fabrication Performance}
\label{tab:hallucination-expanded}
\small
\begin{tabular}{@{}lrrr@{}}
\toprule
\textbf{Model} & \textbf{Grounding} & \textbf{Fabrication} & \textbf{Overall} \\
 & \textbf{(L1-L4)} & \textbf{(L11-L12)} & \textbf{Accuracy} \\
\midrule
GLM-4.5 & 96.7\% & 2.1\% & 97.3\% \\
GLM-4.6 & 93.8\% & 9.7\% & 92.1\% \\
DeepSeek-V3.1 & 95.1\% & 17.3\% & 88.9\% \\
Qwen3-235B & 92.7\% & 15.5\% & 88.6\% \\
Qwen3-235B-FP8 & 92.7\% & 16.5\% & 88.1\% \\
Qwen3-Coder-480B & 90.1\% & 18.4\% & 85.9\% \\
Qwen3-Next-80B & 91.4\% & 19.9\% & 85.7\% \\
Llama-3.1-70B & 86.3\% & 38.8\% & 73.7\% \\
Llama-3.3-70B & 81.4\% & 45.0\% & 68.2\% \\
Granite-4-Small & 67.5\% & 54.5\% & 56.5\% \\
\bottomrule
\end{tabular}
\vspace{0.5em}

\noindent\footnotesize\textit{Grounding measures accuracy on facts present in corpus. Fabrication shows hallucination rate on non-existent facts (lower is better). Results averaged across 4 question sets $\times$ 8 runs = 32 evaluations per model.}
\end{table}

\textbf{Key findings:}

\begin{itemize}
    \item \textbf{GLM-4.5 exhibits exceptional hallucination resistance:} With only 2.1\% fabrication rate, GLM-4.5 almost never claims to find information that does not exist. This is remarkable given that the questions are designed to be plausible (asking about entities that \textit{could} exist but don't).

    \item \textbf{Grounding ability does not predict fabrication resistance:} DeepSeek-V3.1 achieves 95.1\% grounding accuracy (second only to GLM-4.5) but has 17.3\% fabrication rate - 8$\times$ higher than GLM-4.5. Models can be excellent at finding facts that exist while still being prone to ``finding'' facts that don't.

    \item \textbf{Llama models show elevated hallucination rates:} Both Llama-3.1-70B (38.8\%) and Llama-3.3-70B (45.0\%) exhibit substantially higher fabrication rates than similarly-sized models. Notably, the newer Llama-3.3 performs \textit{worse} on hallucination resistance than Llama-3.1.

    \item \textbf{Quantization has minimal impact:} Qwen3-235B and its FP8 quantized variant show nearly identical performance (15.5\% vs 16.5\% fabrication), suggesting that quantization does not significantly affect hallucination behavior for this model. However, not all models and quantization methods are equal. Future work will examine the effects of quantization on enterprise reliability more broadly.
\end{itemize}

\textbf{Within-corpus stability:} Table~\ref{tab:within-corpus-stability} shows fabrication rates across the four independent question sets. Most models exhibit stable behavior: GLM-4.5 shows only 1.4 percentage points spread, while DeepSeek-V3.1 and Qwen3-Coder-480B show 2.0 points each. Notably, Granite-4-Small has the smallest spread (0.7 pts) despite having the worst absolute performance - it consistently hallucinates at the same high rate regardless of which questions are asked.

\begin{table}[ht]
\centering
\caption{Within-Corpus Stability: Fabrication Rate Across Question Sets}
\label{tab:within-corpus-stability}
\small
\begin{tabular}{@{}lrrrrc@{}}
\toprule
\textbf{Model} & \textbf{HA} & \textbf{HB} & \textbf{HC} & \textbf{HD} & \textbf{Spread} \\
\midrule
GLM-4.5 & 2.0\% & 1.2\% & 2.7\% & 2.5\% & 1.4 pts \\
GLM-4.6 & 9.9\% & 7.9\% & 10.7\% & 10.2\% & 2.8 pts \\
DeepSeek-V3.1 & 17.7\% & 18.3\% & 17.0\% & 16.3\% & 2.0 pts \\
Qwen3-235B & 11.2\% & 18.5\% & 16.2\% & 16.0\% & 7.3 pts \\
Qwen3-235B-FP8 & 14.0\% & 19.8\% & 16.5\% & 15.7\% & 5.8 pts \\
Qwen3-Coder-480B & 18.2\% & 19.6\% & 18.2\% & 17.6\% & 2.0 pts \\
Qwen3-Next-80B & 18.0\% & 19.2\% & 20.8\% & 21.7\% & 3.7 pts \\
Llama-3.1-70B & 38.0\% & 37.5\% & 38.5\% & 41.3\% & 3.8 pts \\
Llama-3.3-70B & 42.8\% & 45.1\% & 43.5\% & 48.5\% & 5.7 pts \\
Granite-4-Small & 54.3\% & 54.9\% & 54.2\% & 54.7\% & 0.7 pts \\
\bottomrule
\end{tabular}
\vspace{0.5em}

\noindent\footnotesize\textit{Fabrication rate (\%, lower is better) for each of four independent question sets on the same 128K corpus. Spread = max - min across sets. Each cell averages 8 runs.}
\end{table}

Taken together with the findings in Section~\ref{sec:cross-corpus}, this confirms the feasibility and practicality of RIKER's regenerable corpora and regenerable test set features for contamination resistance. Both approaches show stable measurements, suggesting the measurement of capabilities instead of particular question-specific artifacts.

\section{Discussion}

Our evaluation reveals several findings with implications for both researchers and practitioners deploying LLMs in enterprise contexts. Before discussing specific results, we distinguish between RIKER as an implementation and as a methodology - the latter being the more generalizable contribution.

\subsection{RIKER as Methodology}

RIKER is one implementation of a more general approach: \textit{ground-truth-first synthetic evaluation}. The core methodology consists of three principles:

\begin{enumerate}
    \item \textbf{Ground truth precedes documents.} Define the facts, entities, and relationships first. Generate documents that embed these facts second. This inverts the traditional approach of extracting ground truth from existing documents.
    \item \textbf{Questions derive from ground truth.} Generate questions whose answers are known by construction, not by human annotation. This enables deterministic scoring without reference models or human judges.
    \item \textbf{Regenerability enables contamination resistance.} Because documents and questions are procedurally generated, fresh instances can be created at will. Benchmark validity does not depend on any particular corpus remaining unseen.
\end{enumerate}

The effectiveness of regenerability depends on implementation quality: document distinctiveness requires sufficient randomization of templates, entity pools, and variable combinations. A RIKER-based document generator that lacks sufficient document diversity and distinctiveness due to limited randomization provides little to no protection against contamination.

Beyond contamination resistance, ground-truth-first generation enables unprecedented scale. Traditional benchmarks require human annotation for each test item, making large-scale evaluation expensive. Many recent benchmarks substitute LLM-as-a-judge for human annotation, but this introduces judge model biases, adds inference cost and increases overall evaluation time. RIKER's automatic ground truth and deterministic scoring eliminate both bottlenecks - our 21+ billion token evaluation would be prohibitively expensive with human annotation and impractically slow with LLM judges. Deterministic scoring also ensures perfect reproducibility: the same response always receives the same score.

Our specific implementation to demonstrate the methodology uses commercial leases, sales agent field reports, and HR records - but these are incidental. The methodology applies wherever synthetic documents can be generated from structured ground truth: medical records from patient databases, legal documents from case parameters, scientific papers from experimental results, financial reports from transaction logs.

The stability validations (Sections~\ref{sec:cross-corpus} and \ref{sec:hallucination-analysis}) confirm that this methodology produces reliable measurements. Cross-corpus stability shows that different document instantiations yield consistent rankings. Within-corpus stability shows that different question sets on identical documents yield consistent results. These properties should transfer to any properly implemented ground-truth-first benchmark, not just our specific implementation in RIKER.

\subsection{Cross-Corpus and Within-Corpus Stability}

RIKER's core design hypothesis - that generating documents FROM ground truth rather than extracting ground truth FROM documents enables reliable evaluation - is supported by the stability analyses. Cross-corpus validation (Section~\ref{sec:cross-corpus}) shows that different document instantiations with identical parameters produce consistent measurements: models that perform well on Set A perform well on Sets B, C, and D. Within-corpus validation (Section~\ref{sec:hallucination-analysis}) shows that different question sets on the same corpus also yield stable results. Together, these findings confirm that RIKER measures model capability rather than corpus-specific or question-specific artifacts.

This addresses the fundamental contamination problem identified in our gap analysis. Unlike static benchmarks where performance may reflect memorization of specific examples, RIKER's regenerable corpora create fresh, never-before-seen documents while maintaining the same underlying evaluation parameters. The benchmark can be regenerated indefinitely without losing validity. 

\subsection{Context Length Claims Exceed Usable Capacity}

Models frequently advertise 128K or 200K token context windows, but our results suggest these claims overstate usable capacity for enterprise tasks. Top-tier models achieve over 80\% accuracy at 32K tokens but show meaningful degradation at longer contexts. Some models exhibit catastrophic failures: GLM-4.6 collapses from 70.9\% accuracy at 128K to 26.6\% at 200K, despite nominally supporting both context lengths.

For practitioners, this implies that marketed context length should not be conflated with effective context length. Enterprise deployments requiring long-context processing should validate performance at their actual operating context sizes rather than relying on vendor specifications.

\subsection{Aggregation Reveals a Capability Gap}

Single-document extraction and cross-document aggregation appear to require fundamentally different capabilities. Models that excel at extracting facts from individual documents often struggle when required to aggregate information across multiple documents. This is not simply a matter of ``more work'' - aggregation questions require the model to identify relevant information scattered across the corpus, perform accurate extraction from each source, and combine results correctly.

RIKER's multi-level taxonomy explicitly separates these capabilities, revealing that aggregation accuracy consistently lags extraction accuracy across all models tested. This finding validates the taxonomy design and suggests that benchmarks testing only single-document retrieval may overestimate model capability for realistic enterprise workloads that inherently involve multi-document reasoning.

\subsection{Grounding and Hallucination Resistance Are Distinct}

Perhaps our most striking finding is that strong grounding performance does not predict hallucination resistance. DeepSeek-V3.1 achieves 95.1\% grounding accuracy - second only to GLM-4.5 - yet exhibits a 17.3\% fabrication rate, eight times higher than GLM-4.5's 2.1\%. Models can be excellent at finding information that exists while simultaneously prone to ``finding'' information that does not exist.

This distinction has significant implications for enterprise deployment. A model that reliably extracts correct answers when information is present may still confidently fabricate answers when queried about non-existent entities. Applications requiring high factual reliability must evaluate both capabilities independently.

\subsection{Limitations}

RIKER's current implementation has several limitations that constrain generalizability:

\textbf{Domain scope.} Our evaluation uses enterprise documents (commercial leases, facility field reports, HR records). While chosen for realism, performance on these document types may not transfer to other domains such as scientific literature, legal contracts, or medical records.

\textbf{Language.} All documents and questions are in English. Multilingual performance remains untested.

\textbf{Architecture.} In this study, we evaluate pure context-stuffing - the entire corpus is provided in the prompt. Future work will test retrieval-augmented generation and agentic retrieval patterns.

\textbf{Synthetic realism.} While our Coherent Simulated Universe approach maintains entity consistency across documents, synthetic documents may lack certain characteristics of real-world data: OCR errors, inconsistent formatting, contradictory information across sources, or domain-specific jargon. Models may perform differently on messier real-world corpora. These characteristics can be modeled into RIKER's (or any RIKER-like system's) synthetic data generation logic, but in this current work and results, such characteristics are not included.

\textbf{Model coverage.} Our evaluation covers 33 models available at the time of testing. The rapidly evolving LLM landscape means new models may exhibit different patterns.

\subsection{Future Work}

Several extensions would strengthen RIKER's utility:

\textbf{Additional document types.} Expanding beyond the current three document types to include contracts, technical specifications, financial statements, and other enterprise documents would broaden applicability.

\textbf{Additional test types.} For hallucination detection in particular, more types of hallucination-specific test types can be added that offer a finer-grained view of hallucination failure modes. Instead of single-answer questions (entity name or amount value), questions that expect lists of specific items (e.g., document sources paired with specific retrieved details each) can provide a more comprehensive hallucination-centered metric.

\textbf{RAG and agentic evaluation.} Instrumenting retrieval systems to answer RIKER questions would enable direct comparison between context-stuffing and various RAG approaches on identical ground truth.

\textbf{Multilingual corpora.} Generating documents in multiple languages would enable cross-lingual evaluation.

\textbf{Adversarial robustness.} Introducing deliberate inconsistencies or contradictions in the corpus could test model behavior when ground truth is ambiguous.

\textbf{Continuous benchmarking.} RIKER's regenerable design enables ongoing evaluation as new models emerge, potentially as an automated leaderboard with fresh corpora for each evaluation cycle.

\section{Conclusion}

We presented RIKER, both as a benchmark for enterprise document understanding and as a demonstration of a more general methodology: \textit{ground-truth-first synthetic evaluation}. The core insight - generating documents FROM known ground truth rather than extracting ground truth FROM documents - enables deterministic scoring, contamination resistance through regenerable corpora, and systematic evaluation of capabilities that matter for real-world deployment.

Our evaluation of 33 models using over 21 billion tokens reveals several key findings. Context length claims frequently exceed usable capacity, with significant performance degradation beyond 32K tokens. Cross-document aggregation proves substantially harder than single-document extraction. Grounding ability and hallucination resistance are distinct capabilities that must be evaluated separately.

The stability analyses confirm that the methodology works: both cross-corpus validation (different documents, same parameters) and within-corpus validation (same documents, different questions) produce consistent measurements. This suggests that any properly implemented ground-truth-first benchmark should exhibit similar stability properties, regardless of domain.

Beyond the specific benchmark, we contribute a replicable methodology for constructing contamination-resistant evaluations in any domain where synthetic documents can be generated from structured ground truth.

\section*{Data Availability}
The experiment data, including all of the generated ground truth, document corpora, test sets, and the various model raw results, will be made available at \url{https://docs.kamiwaza.ai/research/datasets}.

\section*{Acknowledgments}

This research was made possible through the generous provision of GPU compute by  Signal65, who provided four servers with 8x AMD MI300X GPUs each for experimental evaluation. We thank Ryan Shrout, Brian Martin, Mitch Lewis, and Russ Fellows for their support. (\url{https://signal65.com/})

\section*{AI Usage Disclosure}

The researchers used the following generative AI services to assist with the manuscript:

\begin{itemize}
    \item Claude Code: Claude Opus 4.5
    \item Gemini 3 Pro Image (Nano Banana Pro)
\end{itemize}

Most of the language in this paper was drafted by generative AI using RIKER project documents and code, plus certain reference material (especially previous work in PICARD) and the raw and summarized RIKER results in CSV form, and then heavily revised, edited and polished by human researchers. 100\% of this document was read and reviewed several times by human researchers.

In addition, all tables were created through generative AI directly using raw source data, and then reviewed by human researchers. Nano Banana Pro was used to generate the RIKER diagram by feeding it the final draft of this paper.

In all cases, final editorial control, technical validation, and intellectual responsibility rest solely with the human authors. The authors take full responsibility for the accuracy and integrity of all content in this manuscript.

\bibliography{references}

\end{document}